\documentclass[twocolumn]{article}

\usepackage[colorlinks,urlcolor=blue,linkcolor=blue,citecolor=blue]{hyperref}

\usepackage{hyperref}
\usepackage{amsmath}
\usepackage{graphicx}
\hypersetup{ 
 colorlinks=true, 
 linkcolor=red, 
 filecolor=blue, 
 citecolor = blue, 
 urlcolor=blue, 
 }

\begin{document}

\title{An Explainable Machine Learning Approach for Energy Forecasting at the Household Level}

\maketitle

\author{

Pauline Béraud\textsuperscript{1}, Margaux Rioux\textsuperscript{1}, Michel Babany\textsuperscript{1}, Philippe de La Chevasnerie\textsuperscript{1} , Damien Theis\textsuperscript{1}, Giacomo Teodori\textsuperscript{1}, Chloé Pinguet\textsuperscript{1}, Romane Rigaud\textsuperscript{1},  François Leclerc\textsuperscript{1}

\textsuperscript{1}\textit{papernest, 157 boulevard Macdonald, 75019 Paris, France}}

\url{https://www.papernest.com}

\begin{abstract} Electricity forecasting has been a recurring research topic, as is it key to finding the right balance between production and consumption. While most papers are focused on the national or regional scale, few are interested in the household level. Desegregated forecast is a common topic in Machine Learning (ML) literature but lacks explainability that household energy forecasts require. This paper specifically targets the challenges of forecasting electricity use at the household level. This paper confronts common Machine Learning algorithms to electricity household forecasts, weighting the pros and the cons, including accuracy and explainability with well-known key metrics. Furthermore, we also confront them in this paper with the business challenges specific to this sector such as explainability or outliers resistance. We introduce a custom decision tree, aiming at providing a fair estimate of the energy consumption, while being explainable and consistent with human intuition. We show that this novel method allows greater explainability without sacrificing much accuracy. The custom tree methodology can be used in various business use cases but is subject to limitations, such as a lack of resilience with outliers.


\end{abstract}

\section{Introduction}  Forecasting is an essential part of the energy industry and even more so in the electricity provider industry. Aggregated energy forecasts are built to predict the future energy use of the whole customer portfolio. They are used to buy in advance the electricity and limit the exposure of the energy provider to the volatile energy market.

They are decisive in order to supply correctly on the electricity market : buy too much and risk reselling at a lower price than the purchase price ; or buy too little and be at the mercy of last minute price.

\vspace{1cm} 

Household predictions are used to define customers' monthly payments. They have an impact on acquisition, churn, payment delinquency and cash flow projection. These forecasts are used by many teams of an energy supplier company, with various levels of familiarity with mathematical models, particularly machine learning models. In France, most consumers pay a fixed monthly installment for their electricity and are subject yearly to a regularization. In some cases, this can lead to what is known as “bill shock”, the unpleasant and undesired effect of customers having to pay a large bill when their consumption was incorrectly estimated at subscription. Bill shock results in dissatisfaction or even contract termination. Building a precise installments schedule is crucial to avoid this. It is preferable to be able to explain to the customer the amount of their monthly installment which requires the prediction models to be accessible. 

Moreover, monthly installments have to meet certain requirements in terms of transparency and availability. These requirements are defined by the French Commission de Régulation de l'Énergie (CRE), the independent administrative authority in charge of the smooth operation of the electricity and gas markets.

The stakes are both financial and customer-focused: reduce terminations, build a relationship of trust with customers, and have an accurate view of portfolio consumption in the short, medium, and long term. That is why this issue is key to making strategic decisions.

\vspace{1cm} 

Papers focusing on household predictions highlights the difference on the data type and source needed between micro and macro level. [1,2,3] Previous studies settles on using smart meters readings to measure household electricity consumption.[4] That data is typically unavailable when people move to a new home which makes it unsuitable to our problem.

Given the diversity of customers and their varying consumption patterns, accurately estimating their consumption at the time of subscription presents numerous challenges, while still requiring the models to be logical and explainable. Machine learning literature [5,6] describes various algorithms and strategies that seem adapted to forecasting but most prediction algorithms applied to the energy sector are typically used at regional or national scales. [1,2] 

\vspace{1cm} 

Tree-based models (TBM) and linear models are among the most interpretable approaches [7]; however, TBMs are primarily used for population-level predictions, and linear models often lack precision and accuracy. Probabilistic forecasting methods are recognized for their reliability, flexibility, and scalability, making them particularly suitable for household load forecasting. Deep Learning models and Gradient Boost offer strong performance but are too much of a "black box" for monthly billing use cases and related business decisions. [3] For this reason, we did not include Deep Learning models in this paper but included Gradient Boost as a performant candidate to be benchmarked with more transparent models.
From a mathematical and modeling perspective, the main challenges lie in the presence of a significant number of outliers, which should not affect the results for customers with a typical profile. Outliers are not rare as they can correspond to a second home but our objective is to predict the "median" behavior rather than the average one. In addition, to keep results explainable, an important criteria is that the outputs remain consistent and align with business expectations. For instance, a dwelling with more occupants is expected to consume more energy, while a dwelling with a higher meter capacity is expected to have higher energy consumption.

The aim of this paper is to benchmark traditional Machine Learning methods, highlighting their advantages and disadvantages. It also introduces a new approach based on a customizable decision tree that addresses both the issues and challenges of consistency and explainability.

\section{Methodology} 

We aim at evaluating and testing various models with different levels of algorithmic complexity. We want to identify which type of algorithm is most relevant, both in terms of performance and business logic, as well as transparency. Striking a balance between the availability of data at the time of consumption calculation—when the customer signs up—along with the accuracy and consistency of the results, is central to this analysis. 

Several algorithms have been compared in this study, each offering its own advantages and disadvantages :

\begin{itemize}
    \item \textbf{Legacy algorithm}: A basic decision tree based on housing information, which tends to overestimate consumption. It was the prior method to beat.
    \item \textbf{Gradient Boost algorithm}: Highly accurate but lacks explainability.
    \item \textbf{Random Forest Regressor}: Offers high accuracy, explainability, and consistency.
    \item \textbf{Linear Regression}: Simple, but less accurate.
    \item \textbf{Custom Decision Tree}: Transparent, adjustable, and business-friendly.
\end{itemize}

Table 1 below provides an overview of the expected strengths and drawbacks of each method:

\begin{figure}[h!] 
    \centering
    \includegraphics[width=0.4\textwidth]{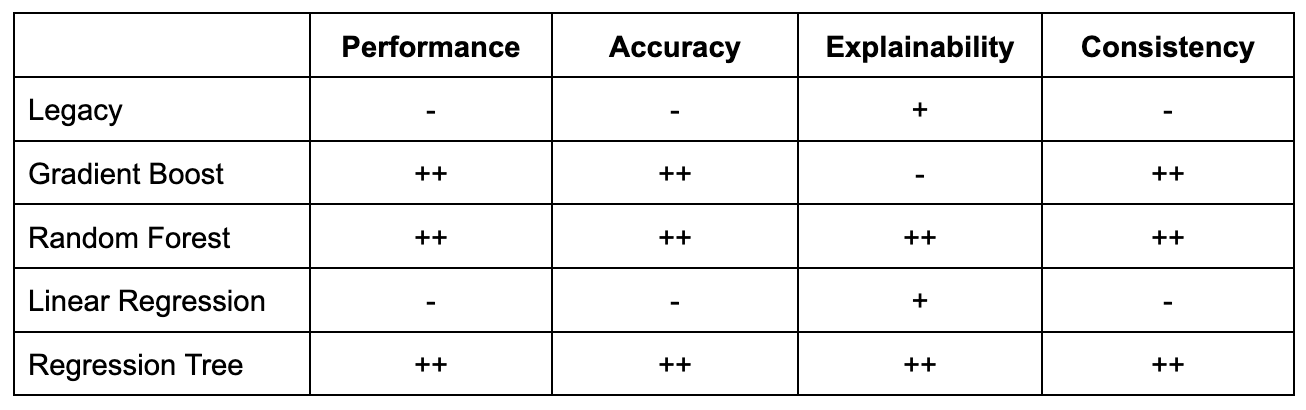}
    \caption{Strengths and drawbacks of each method}
\end{figure}

To compare these five algorithms and ensure reliable results, we required historical data from a portfolio of users with a sufficiently long consumption measurement period to capture enough data, particularly to account for seasonal patterns and behaviors specific to certain periods (winter vs. summer).

\vspace{1cm} 

The aim is to estimate the yearly electricity consumption, known in French as the "consommation annuelle de référence" (CAR), expressed in kWh.
Given that papernest is a new electricity provider and most customers are recent, we determined that 70 days of electricity consumption serves as a good proxy for yearly behavior. Thus, our dataset is composed of papernest consumer data with at least 70 days of consumption readings, most of them acquired through \url{https://www.fournisseur-energie.com}. This is accessible because we automatically collect our customers’ meter consumption every 30 days. Additionally, we only considered consumers who provided all the necessary information for the study. This information primarily pertains to housing data, along with some user-specific data, although we later found the latter to be of lesser importance. These parameters are listed in the appendix.

\vspace{1cm} 

One third of the sample was used for testing, while the remaining two thirds were used for training the machine learning algorithms. The first step was to analyze which parameters had the greatest impact on the estimated consumption. Housing data served as the starting point, and this was confirmed by the Gradient Boosting and Random Forest algorithms.

\vspace{1cm} 

Figure 2 displays features importance according to the algorithm :

\begin{figure}[h!] 
    \centering
    \includegraphics[width=0.5\textwidth]{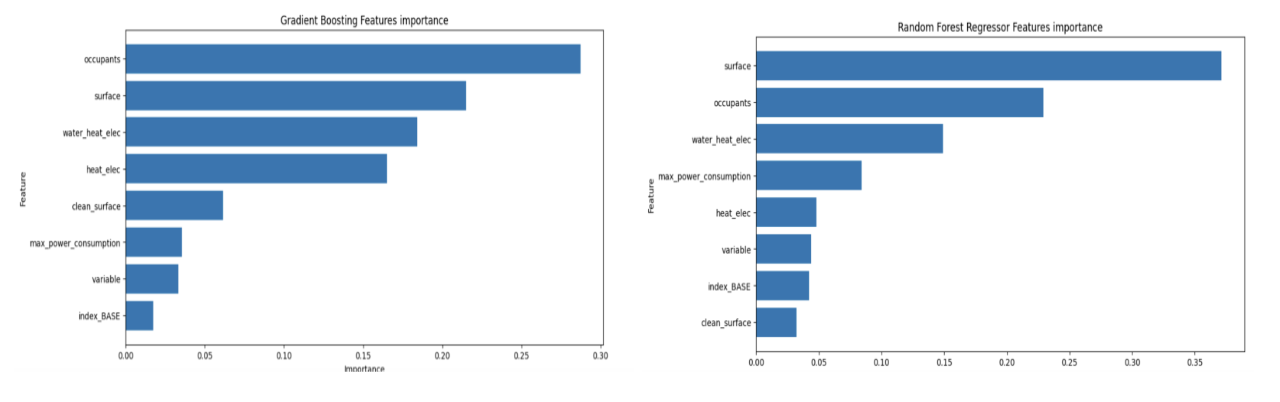}
    \caption{Features importance for Gradient Boosting and Random Forest Regressor models}
\end{figure}

It was clear and expected that the number of occupants, the surface area, and the type of heating and water heating are the most important parameters. These will be used to construct our own decision tree.

While the Gradient Boosting and Random Forest algorithms provide valuable insights into the most significant features, they remain "black boxes" at the individual level, as there is limited transparency regarding these results, and it’s not always easy to understand or explain the underlying business logic. Although SHAP values could help clarify how each feature impacts each prediction, the results are unstable across different users. Since the goal is to create a tool that is accessible and usable by a wide range of stakeholders, this approach did not fully meet our needs.

\vspace{1cm} 

We then shifted our focus to tree-based models, which are known for their precision and robustness [7]. Our objective was to develop a custom decision tree, designed as a typical decision tree with a minimum leaf size and standard splitting metrics. To achieve this, we conducted a grid search to build the tree. The split was binary, and the same variable could be used multiple times if necessary, depending on its importance, but we chose to limit the tree depth to seven to ensure logical consistency and reliability.

The prior analysis using Gradient Boosting and Random Forest, with their insights on feature importance, helped narrow down the number of candidate variables to meet our objectives. We ensured that these variables aligned with business expectations to maintain a logic that could be easily explained. Historical data from papernest clients also enabled us to identify typical profiles of low-consumption customers, which are a combination of the parameters identified in the earlier grid search. These profiles were valuable inputs for building our decision tree and refining our understanding of consumer trends.

\vspace{1cm} 

These findings also guided our final decision on the depth and complexity of the regression tree. We observed that a tree with too great a depth resulted in very small buckets, which is suboptimal from both a performance and business logic standpoint. For this reason, we defined a minimum bucket size to ensure consistent buckets, while also aiming for a constant depth with splits occurring at the same level for the same feature, ensuring better explainability.

\vspace{1cm} 

Figure 3 displays a visualization of a possible tree architecture : 

\begin{figure}[h] 
    \centering
    \includegraphics[width=0.5\textwidth]{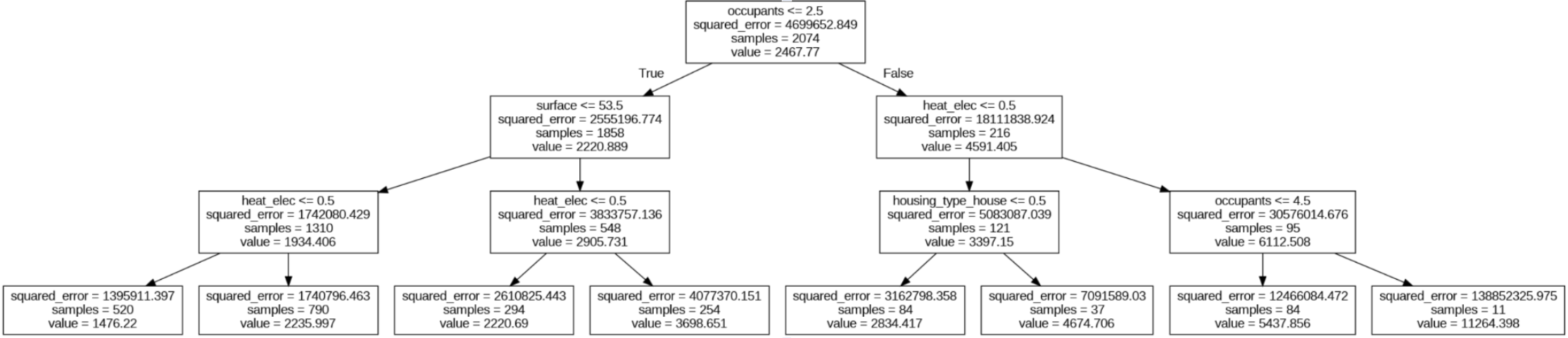}
    \caption{Possible decision tree architecture}
\end{figure}

Once all these assumptions had been made and the tree constructed from scratch, it was possible to compare these different algorithms using common metrics from a purely mathematical point of view. 

\section{Results}

The parameters finally chosen for the custom decision tree were : 
\begin{itemize}
    \item Low consumption boolean (combination of housing parameters to define a low-consumer profile)
    \item Index (Base or peak hours/off-peak hours)
    \item Number of occupants
    \item Type of heating
    \item Type of water heating
    \item Surface
    \item Surface
\end{itemize}

We used the surface feature twice to increase the number of splits and achieve four splits even if the tree is binary.

\vspace{1cm} 

After conducting user interviews with our customers support team, we found that consumers didn’t fully understand why they were assigned the same estimated consumption when they adjusted the surface of their home. As a result, we decided to linearize the predicted CAR by introducing a parameter linked to the surface area, allowing for an infinite number of CAR values, rather than just those proposed by the initial tree. This adjustment brought the estimates closer to expectations, with negligible impact compared to the decision tree splits.

\vspace{1cm} 

We began by calculating the root-mean-square deviation (RMSD), the median absolute deviation (MAD), and the mean absolute error (MAE) for each algorithm, as these metrics are commonly used to assess model performance.

\[
\text{RMSD} = \sqrt{\frac{1}{n} \sum_{i=1}^{n} (y_i - \hat{y}_i)^2}
\]

\begin{itemize}
    \item \( n \) : total number of data points
    \item \( y_i \) : true value
    \item \( \hat{y}_i \) : prediction
\end{itemize}

\[
\text{MAD} = \text{median} \left( |y_i - \text{median}(y)| \right) \quad \text{for } i = 1, 2, \ldots, n
\]

\begin{itemize}
    \item \( n \) : total number of data points
    \item \( y_i \) : each value
    \item \( \text{median}(y) \) : median value
\end{itemize}

\[
\text{MAE} = \frac{1}{n} \sum_{i=1}^{n} |y_i - \hat{y}_i|
\]

\begin{itemize}
    \item \( n \) : total number of data points
    \item \( y_i \) : true value
    \item \( \hat{y}_i \) : prediction
\end{itemize}

We had a total sample size of 42 000 papernest clients, two third used for training and one third for testing.

Table 4 displays results for each model, comparing the main metrics mentioned above :

\begin{figure}[h] 
    \centering
    \includegraphics[scale=0.3]{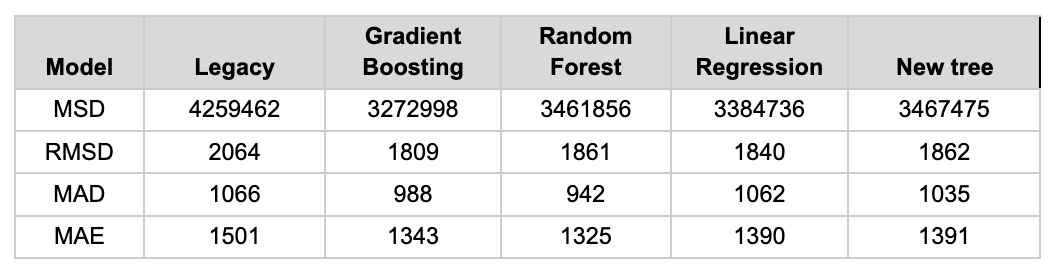}
    \caption{Key metrics for each model}
\end{figure}

Figure 5 displays the distribution of the gaps between the estimated CAR, derived from meter readings and used as the target, and the CAR calculated by the various algorithms. 

\begin{figure}[h] 
    \centering
    \includegraphics[scale=0.4]{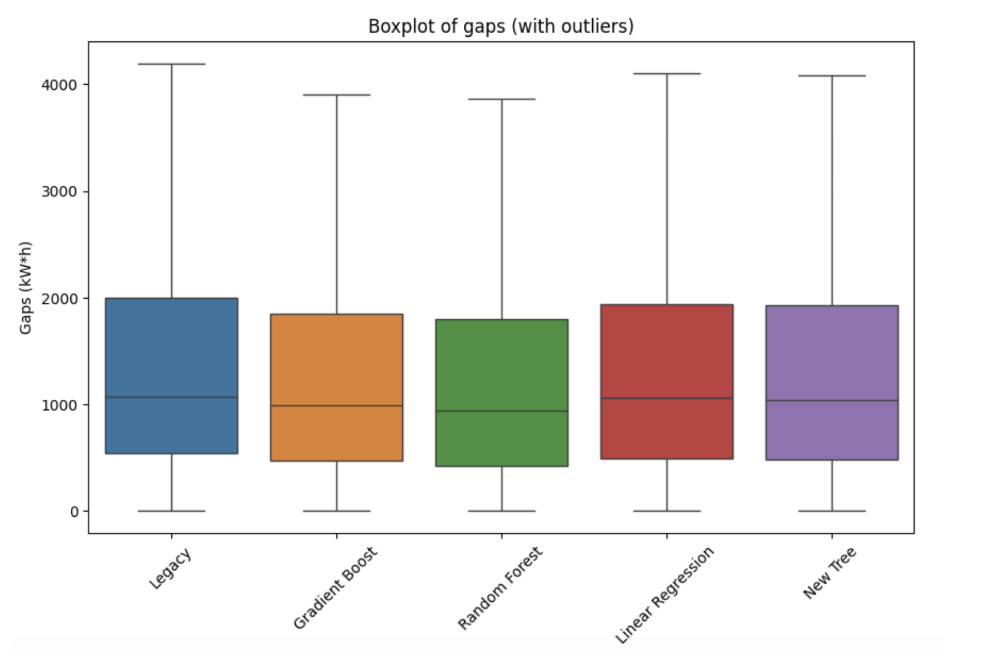}
    \caption{Gaps distribution for each model in absolute value}
\end{figure}

Analysis revealed that our customer tree performs poorly on customers with unusual consumption. People with CAR below 1000 kWh are generally second home who cannot be predicted with the data on hand. High consumption above 10000 kWh was also excluded. We retrained all the models without outliers but kept outliers in the test set.

\vspace{1cm} 

Table 6 displays a clear improvement of the various metrics previously calculated, even if outliers remain in the test set. It also shows how that the different models are not equally outliers resilient.

\begin{figure}[h] 
    \centering
    \includegraphics[scale=0.3]{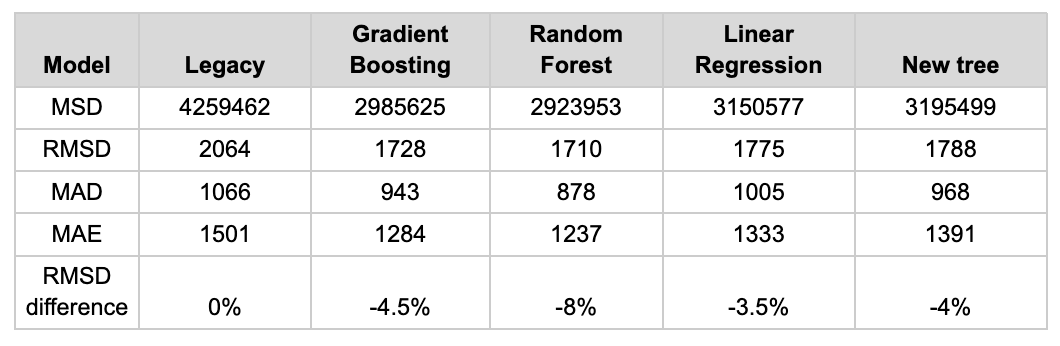}
    \caption{Key metrics for each model without outliers}
\end{figure}

Figure 7 displays the distribution of gaps in absolute value between the estimated car and the car calculated by the various algorithms, to see if these improvements were confirmed. 

\begin{figure}[h] 
    \centering
    \includegraphics[scale=0.4]{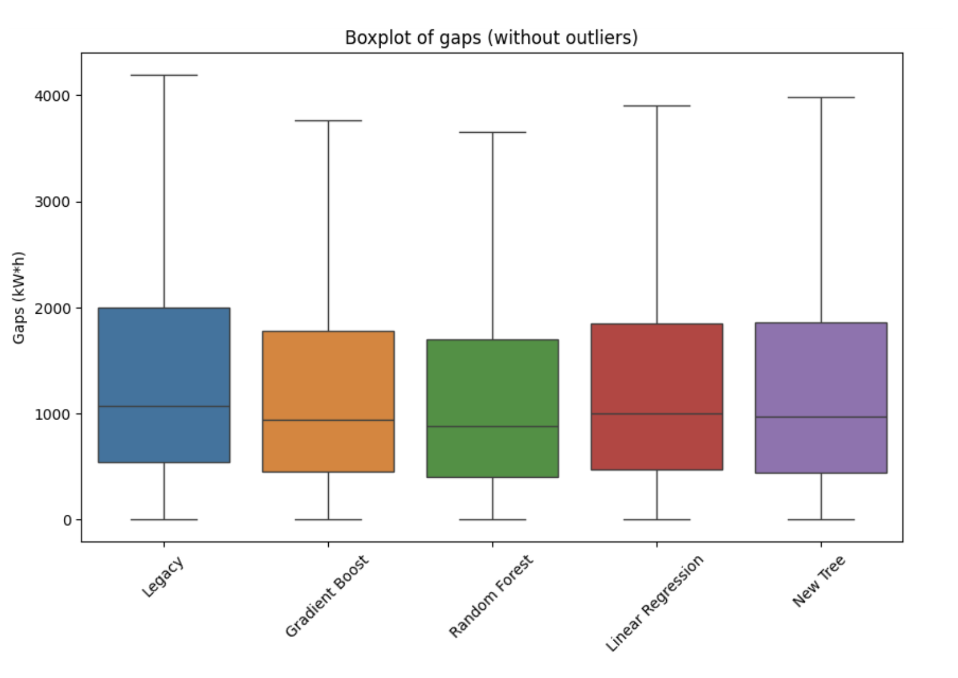}
    \caption{Gaps distribution for each model without outliers in absolute value}
\end{figure}

\vspace{2cm} 

This box plot, in addition to table 6, confirms that the Random Forest algorithm is the best performer. The custom tree is a clear improvement over the legacy and stands the comparison with other models.

\vspace{1cm} 

Figure 8 displays the gap distribution between the estimated car and the car calculated by the various algorithms in relative value:

\begin{figure}[h] 
    \centering
    \includegraphics[scale=0.4]{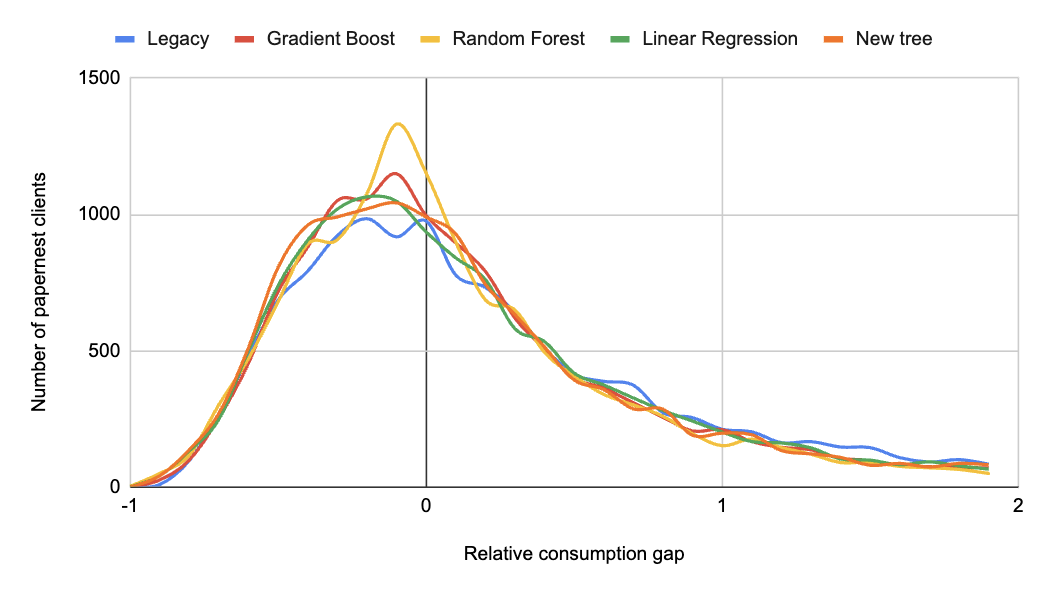}
    \caption{Gaps distribution for each model without outliers in relative value}
\end{figure}

For example, for the Random Forest model, we have almost 500 users with a positive gap around 40\% and more than 1300 users with a negative gap around 10\%. The graphs shows that the gap distribution of the customer tree is one of the best among the models tested.

Since this analysis was also intended for business and finance-oriented teams, we needed to translate these results into financial terms to assess the value generated by each model.

\vspace{1cm} 

Figure 9 displays the previous gaps converted in price with the estimate that 1kW*h = 0.2516€ (French regulated electricity price as of February 1st 2024) :

\begin{figure}[h] 
    \centering
    \includegraphics[scale=0.4]{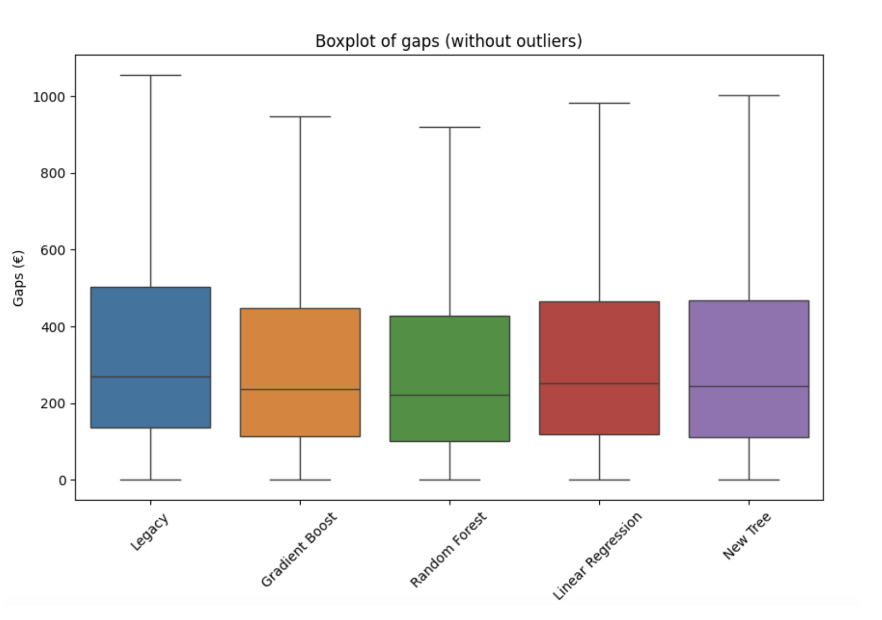}
    \caption{Gaps distribution for each model without outliers in price}
\end{figure}

We have developed an explainable decision tree that performs similarly to conventional machine learning models. In practice, updating the monthly installments before the first anniversary is important to prevent “bill shock" from happening to papernest customers.

The advantage of this customized tree is that it's easy to see the weight of each feature, and therefore to understand the different steps and splits in the tree that have led to this specific CAR value. It also opens the door to improvements thanks to its high level of explainability.

\section{Conclusion}

We chose to build and implement a logical and explainable decision tree that is almost as accurate as other machine learning algorithms. It achieves our objective to be both explainable and performant. However, we found that this model is not highly resilient to outliers. In this analysis, we opted to exclude outliers based on the different results, even though we acknowledge their existence. A potential improvement would be to build a 2-steps model : the first one would be to predict outliers and the second one to predict consumption.

\vspace{1cm} 

Another avenue for future exploration could be to investigate other types of algorithms, such as Deep Learning, despite their main drawback of lacking transparency and being complex to interpret.[3]

\vspace{1cm} 

We could also define alternative metrics beyond the commonly used ones, which primarily compare actual consumption to forecasts. For example, we could employ pairwise comparisons to challenge explainability. We could also look at metrics related to churn to examine the correlation between consumption gaps at subscription and churn rates. 
These metrics would more effectively capture the strengths and weaknesses of the different forecasting models, focusing on user characteristics and behaviors to identify inconsistencies and validate business logic.

\vspace{1cm} 

The main value of this paper lies in our effort to combine precision and the constraints of the sector, along with its specific features, in both the methodology used and the tools tested, seeking accuracy without sacrificing consistency. Furthermore, external factors such as energy prices, new regulations, political events, and temperature can also influence outcomes, adding complexity and not necessarily leading to greater efficiency.[8] Housing-related ones such as energy performance diagnoses have also a great impact on results.

\vspace{1cm} 

Ultimately, we recognize that calculating annual consumption is just the first step toward gaining a precise and comprehensive understanding of a portfolio of users' consumption. Once this initial output is determined, the next challenge is to project it accurately based on seasonality, consumption habits, and various other external factors.

\subsection{Acronyms and Abbreviations}

\begin{itemize}
    \item \textbf{CAR}: Consommation Annuelle de Référence
    \item \textbf{CRE}: Commission de Régulation de l'Énergie
    \item \textbf{MAD} : Median Absolute Deviation
    \item \textbf{MAE} : Mean Absolute Error
    \item \textbf{ML}: Machine Learning
    \item \textbf{RMSD}: Root-mean-square deviation
    \item \textbf{TBM}: Tree-based Model
\end{itemize}

\subsection{List of variables}

\begin{itemize}
    \item Surface
    \item Heating type
    \item Water heating type
    \item Cooking type
    \item Number of occupants
    \item House type
    \item Index (Base or peak hour/off-peak hour)
    \item Maximum power consumption
    \item Number of meter reading days
\end{itemize}

\end{document}